\documentclass[final,5p,times]{elsarticle}
\usepackage{graphicx}
\usepackage{subcaption}
\usepackage{epsfig}
\usepackage{siunitx}
\usepackage{booktabs}
\usepackage[T1]{fontenc}
\usepackage{multirow} 
\usepackage{tabularx}
\usepackage{array}
\usepackage[namelimits]{amsmath} 
\usepackage{amssymb} 
\usepackage{amsmath}            
\usepackage{amsfonts}            
\usepackage{mathrsfs} 
\usepackage{float}
\usepackage{cases}
\usepackage{wrapfig}
\usepackage{hyphenat}
\usepackage{multicol}
\usepackage{caption}
\usepackage[labelfont=bf, textfont=normalfont, justification=raggedright, singlelinecheck=false]{caption}
\usepackage{graphicx}
\usepackage{subcaption}
\usepackage{tabularx}  
\usepackage{amsmath}

\usepackage{amssymb}

\journal{arxiv}
\date{}

\begin{document}
\begin{frontmatter}



\title{An Error-Matching Exclusion Method for Accelerating Visual SLAM \tnoteref{13}}


\tnotetext[13]{This work was supported in part by the National Natural Science Foundation of China (No. 62172190), National Key Research and Development Program(No. 2023YFC3805901), the "Double Creation" Plan of Jiangsu Province (Certificate: JSSCRC2021532) and the "Taihu Talent-Innovative Leading Talent" Plan of Wuxi City(Certificate Date: 202110). }
\author[label1]{Shaojie Zhang}
\ead{7213107006@stu.jiangnan.edu.cn}
\author[label1,label2]{Yinghui Wang\corref{cor1}}
\ead{wangyh@jiangnan.edu.cn}
\author[label1]{Jiaxing Ma}
\ead{2458098051@qq.com}
\author[label1]{Wei Li}
\ead{cs_weili@jiangnan.edu.cn}
\author[label1]{Jinlong Yang}
\ead{yjlgedeng@163.com}
\author[label1]{Tao Yan}
\ead{yantao.ustc@gmail.com}
\author[label1]{Yukai Wang}
\ead{ericwangyk22@163.com}
\author[label3]{Liangyi Huang}
\ead{lhuan139@asu.edu}
\author[label4]{Mingfeng Wang}
\ead{mingfeng.wang@brunel.ac.uk}
\author[label5]{Ibragim R. Atadjanov}
\ead{ibragim.atadjanov@gmail.com}
\cortext[cor1]{Corresponding author}
\affiliation[label1]{organization={ School of Artificial Intelligence and Computer Science, Jiangnan University},
            addressline={1800 Li Lake Avenue},
            city={wuxi},
            postcode={214122},
            state={Jiangsu},
            country={PR China}}
\affiliation[label2]{organization={ Engineering Research Center of Intelligent Technology for Healthcare, Ministry of Education},
            addressline={1800 Li Lake Avenue},
            city={wuxi},
            postcode={214122},
            state={Jiangsu},
            country={PR China}}
 \affiliation[label3]{organization={School of Computing and Augmented Intelligence, Arizona State University},
            addressline={1151 S Forest Ave},
            city={Tempe},
            postcode={8528},
            state={AZ},
            country={U.S}}
\affiliation[label4]{organization={Department of Mechanical and Aerospace Engineering, Brunel University},
            addressline={Kingston Lane},
            city={London},
            postcode={UB8 3PH},
            state={Middlesex},
            country={U.K}}    
\affiliation[label5]{organization={Tashkent University of Information Technologies named after al-Khwarizmi},
			addressline={ 108 Amir Temur Avenue},
			city={Tashkent},
		    postcode={100084},
			state={},
		    country={Uzbekistan}}

\begin{abstract}
In Visual SLAM, achieving accurate feature matching consumes a significant amount of time, severely impacting the real-time performance of the system. This paper proposes an accelerated method for Visual SLAM by integrating GMS (Grid-based Motion Statistics) with RANSAC (Random Sample Consensus) for the removal of mismatched features. The approach first utilizes the GMS algorithm to estimate the quantity of matched pairs within the neighborhood and ranks the matches based on their confidence. Subsequently, the Random Sample Consensus (RANSAC) algorithm is employed to further eliminate mismatched features. To address the time-consuming issue of randomly selecting all matched pairs, this method transforms it into the problem of prioritizing sample selection from high-confidence matches. This enables the iterative solution of the optimal model. Experimental results demonstrate that the proposed method achieves a comparable accuracy to the original GMS-RANSAC while reducing the average runtime by 24.13\% on the KITTI, TUM desk, and TUM doll datasets.\end{abstract}



\end{frontmatter}


\section{Introduction}
Real-time capability is a core characteristic of Simultaneous Localization and Mapping (SLAM), garnering increased attention, particularly in visual SLAM \cite{ref19}. Feature point extraction and matching represent the most crucial and time-consuming steps in visual SLAM \cite{ref1}-\cite{ref2}. These steps involve establishing reliable correspondences between images of the same scene captured at different times, from different perspectives, or using different sensors. This process aims to describe relationships between images, or between images and maps, providing essential support for subsequent SLAM tasks such as camera pose estimation, optimization operations, and various mission objectives \cite{ref3}. Due to the local nature of image features, mismatches are prevalent. While strategies like iterative optimization have mitigated error rates to some extent \cite{ref4}, the introduction of matching errors and additional processes adds extra time overhead, exacerbating the challenge of enhancing real-time performance in visual SLAM and becoming one of the bottlenecks in SLAM development.

In recent years, Random Sample Consensus (RANSAC) technology \cite{ref5} has gained preference in visual SLAM for addressing the problem of eliminating mismatched features and has become a widely accepted and versatile approach. However, when dealing with large datasets, achieving the required precision necessitates an increase in the number of iterations, leading to a rapid growth in matching time. This is intolerable for real-time SLAM systems \cite{ref6}. Conversely, reducing the number of matches will compromise system accuracy, thereby reducing the applicability of SLAM. Typically, SLAM often faces the challenge of dealing with large datasets in vast environments, making it difficult to ensure both short matching times and high accuracy simultaneously. To address this, Zhang et al. \cite{ref7} recently introduced the GMS-RANSAC method \cite{ref14} and applied it to SLAM systems to balance this trade-off. The approach involves an initial filtering using the Grid-based Motion Statistics (GMS) algorithm \cite{ref8}, followed by further error-matching elimination using RANSAC. Despite being the most accurate method for visual SLAM in complex environments, GMS-RANSAC still struggles to meet real-time requirements in many scenarios.

Inspired by GMS-RANSAC, this paper proposes an improved method that combines Grid-based Motion Statistics (GMS) and RANSAC for error-matching elimination. By optimizing RANS\\AC's input samples, this method ensures accuracy while significantly reducing the number of iterations, thereby shortening the matching time. Through experimental comparisons, an average time saving of 32\% was achieved. The innovations of this paper can be summarized as follows:

(1)	Introducing a feasibility-based optimization method for GMS results, ensuring the accuracy of samples provided to RAN\\SAC.

(2)	Implementing a strategy to reduce RANSAC input samples, shortening the time for error-matching elimination in the GMS-RANSAC algorithm.

\begin{figure*}[t]
    \centering
    \includegraphics[height=2.5in]{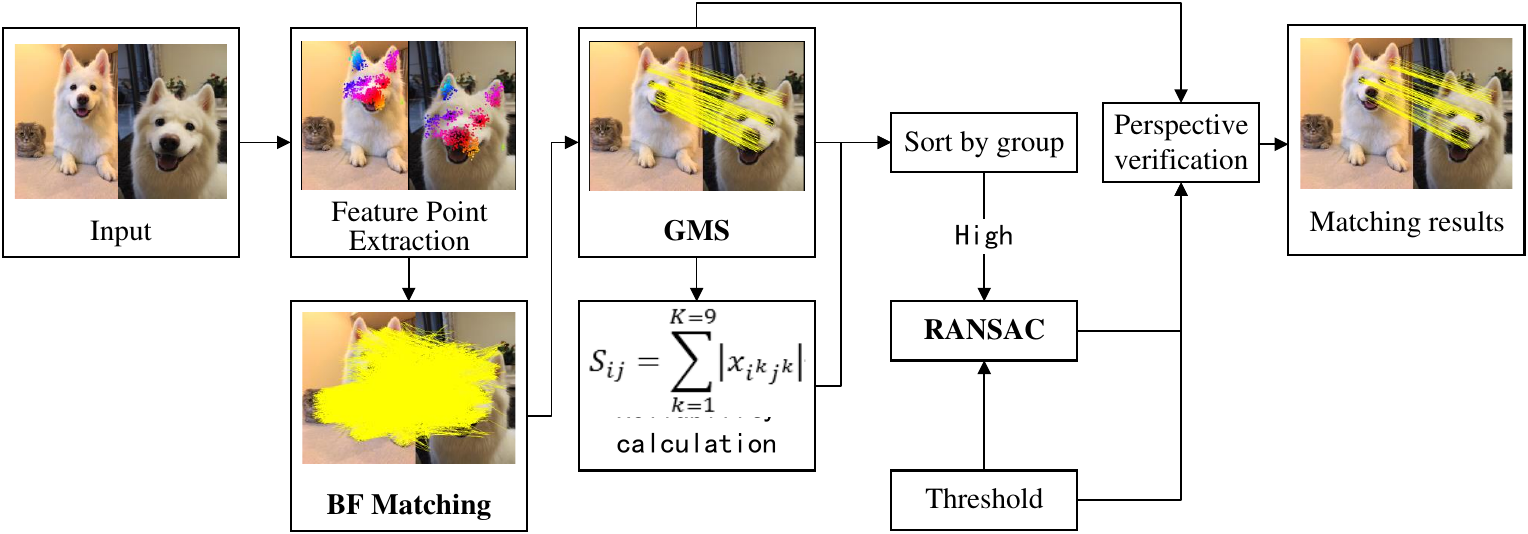}
\caption{Methodological Framework.}
    \label{Fig1}
\end{figure*}

\section{RELATED WORK}
After extracting features in visual SLAM, feature matching can yield a series of preliminary matching results. However, due to repetitive textures, changes in lighting conditions, and camera imaging transformations, a large number of erroneous matches inevitably occur in the matching results. These erroneous matches bring additional time overhead; moreover, the matching filtering phase, which involves removing these erroneous matches from the pairs of matched points generated by feature matching, also consumes additional time resources. Fischler et al. proposed the RANSAC algorithm as early as 1982 [5], which has been widely recognized as the most universally applicable method for erroneous match filtering. It works by randomly selecting samples from the input and computing the best model, with samples that fit the model considered inliers, i.e., correct matches. In recent years, to improve the efficiency and accuracy of the RANSAC algorithm, many researchers have proposed various improvements. In 2004, Matas et al. \cite{ref9} proposed the Randomized-Random Sample Consensus (R-RANSAC) algorithm, which first verifies a model with N randomly selected sample points, and upon verification, uses all M data points for further validation. If the data in the sample do not fit the model, it is directly abandoned, thereby reducing the computational load and improving operational efficiency. Chum et al. \cite{ref10} in 2005 introduced the PROSAC (Progressive Sampling Consensus) algorithm, which, unlike the RANSAC method that uniformly samples from the entire set, samples from an increasingly larger set of best corresponding points to save computational effort and improve running speed. In 2012, Xu et al. \cite{ref11} proposed selecting random subsets of input data based on statistical ideas and fitting the best model from all choice points in each subset, leading to an exponential decrease in the number of points used and thereby improving the efficiency of the algorithm. Ci et al. \cite{ref12} in 2022 suggested using the top 20\% and the remaining 80\% of all matching pairs to form a new sample set. They calculated the matrix model with the first 20\% of matching pairs and validated the computed model, using the transformation model to verify whether the remaining 80\% were inliers. This improved the proportion of inliers in the sample set, effectively enhancing the accuracy and speed of image matching, but the fixed nature of the 20/80 split reduced its adaptability. Bian et al. in 2017 proposed the GMS (Grid-based Motion Statistics) algorithm \cite{ref8}, using statistical methods to count the number of matches in local grid areas and determine whether all matches in that area are correct based on the count. In 2018, Chen et al. \cite{ref13} optimized GMS, changing the nine-grid statistical method to a five-grid one and reducing the computation of seven rotation matrices to three, though this reduced the algorithm's rotational invariance. Zhu et al. \cite{ref14} in 2019 introduced GMS-RANSAC based on improved grid motion statistical features, incorporating the principle of distance consistency to eliminate outliers, increasing accuracy but also running time. Lan et al. \cite{ref18} in 2020 proposed an improved version of GMS-RANSAC, using bidirectional brute-force matching and the GMS algorithm, followed by further outlier removal based on RANSAC, and finally using the GMS algorithm to verify complete feature pair matching. In 2022, Zhang et al. \cite{ref7} proposed the application of GMS-RANSAC in ORB-SLAM2, currently the most effective method for outlier elimination in terms of accuracy, although setting a threshold to reduce the number of iterations decreases the running time at the cost of precision, especially in large-scale data where the trade-off between accuracy and speed is irreconcilable.

Specifically, the GMS-RANSAC algorithm involves using the RANSAC algorithm to filter the results obtained from the GMS algorithm. However, in the process of solving for the optimal model in RANSAC, it still requires randomly selecting all samples for iterative solving, which is time-consuming. To address this, this paper proposes using the credibility obtained from the GMS algorithm as a preliminary judgment condition for the RANSAC algorithm. The match pairs are sorted and grouped based on their credibility, and the RANSAC algorithm will preferentially select from match pairs with higher credibility, thus quickly obtaining the optimal model. This approach significantly reduces the number of iterations while maintaining accuracy, speeding up the convergence of the function and ensuring the real-time requirements of visual SLAM.

\section{METHODOLOGY}
The technical approach of the method proposed in this paper is illustrated in Figure 1. Initially, feature points are extracted from two images using the ORB algorithm \cite{ref15}. Subsequently, the Hamming distance is computed based on descriptors to perform brute-force matching of ORB features. Following this, the obtained matching results undergo a coarse screening using the GMS algorithm to reduce the number of matches. The GMS algorithm evaluates the correctness of feature pairs by considering the quantity of matching pairs in the surrounding neighborhood of the feature points. Adequate matching pairs in the vicinity indicate a correct match, while almost zero matching pairs suggest an incorrect match. To quantify this, the paper introduces the concept of confidence, denoted as the number of matching pairs in the neighboring area of a feature point. The higher the number of matching pairs, the greater the confidence, indicating a higher likelihood of the current feature point being a correct match. In summary, the method sorts and groups matching pairs based on the confidence obtained from the GMS algorithm. Additionally, the RANSAC algorithm's original random selection is modified to prioritize selecting from matching pairs with higher confidence levels, continuing until the desired precision is achieved. This modification aims to reduce the number of iterations and improve overall runtime efficiency.

\subsection{Analysis of GMS-RANSAC Method}
GMS, known as the Grid-based Motion Statistics method, is a fast and robust feature matching approach. It relies on the principle that feature points in areas of consistent motion are likely to have a higher number of matching points nearby. This method assesses the correctness of a match by counting the number of matching points in the surrounding area. Correctly matched feature points will have a sufficient number of matching pairs nearby, while incorrectly matched feature points will have almost no matching pairs around them.

RANSAC, also known as Random Sample Consensus, uses an iterative method to estimate the parameters of a mathematical model from a set of observed data that contains outliers. The RANSAC algorithm assumes that the data contains both correct (inliers) and incorrect (outliers) data points. In the illustrative diagram of RANSAC, the blue dots represent inliers, the red dots represent outliers, and the line represents the model line found by RANSAC.
\begin{figure}[H]
    \centering
    \includegraphics[height=1.7in]{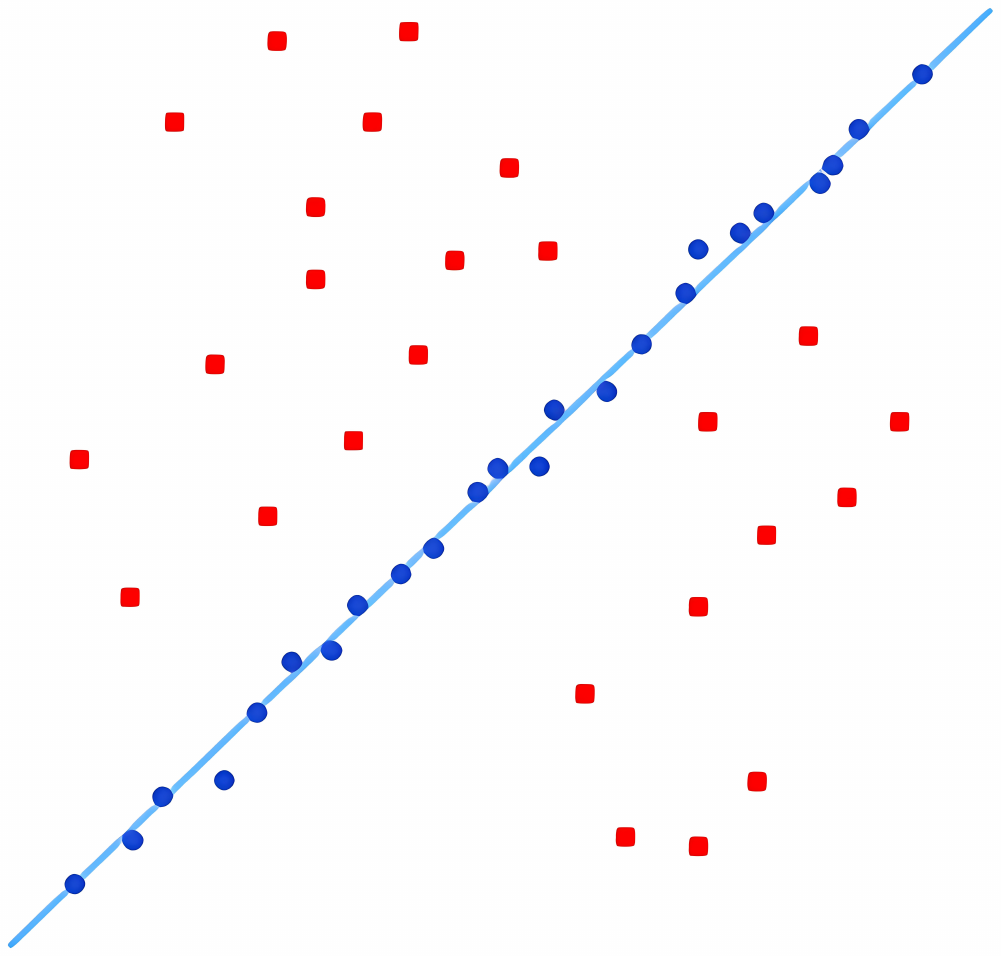}
\caption{RANSAC schematic diagram.}
    \label{Fig2}
\end{figure}
According to the RANSAC algorithm, the estimation of the number of iterations is as follows: 

Assuming the probability of an inlier in the dataset is t, as shown in Equation (1):
\begin{equation}
	\label{eq1}
	t=\frac{n_{inliers}}{n_{inliers}+n_{outliers}}
\end{equation}
Here, $n_{inliers}$ is the number of inliers, and $n_{inliers}$ is the number of outliers.

The probability that all n points are inliers in a single iteration is $t^{n}$. Therefore, the probability of at least one point being an outlier (sampling failure) in n points is  $1-t^{n}$. In other words, after k iterations,  $(1-t^{n})^{k}$ represents the probability that none of the $k$ random samples contains all inliers. Hence, the probability p of successfully sampling n correct points to compute the correct model is given by Equation (2).

\begin{equation}
	\label{eq2}
	p=1-(1-t^{n})^{k}
\end{equation}

This leads to the estimation of the number of iterations k as shown in Equation (3):

\begin{equation}
	\label{eq3}
	3k=\frac{log(1-p)}{log(1-t^{n})}
\end{equation}

The optimal model is constrained by the number of iterations, and there is a strong consistency between the upper limit of iterations and the probability of obtaining the best model. Increasing the iteration count increases the probability of obtaining the best model but also reduces the algorithm's speed.

In the ORB-SLAM3 system, RANSAC is iterated fewer times to ensure real-time responsiveness, potentially sacrificing accuracy. The GMS-RANSAC algorithm first uses the GMS algorithm for initial filtering, and then applies the RANSAC algorithm to further refine the results. This approach reduces the number of iterations needed to find the optimal model, thus improving the algorithm's speed. However, it still requires random sampling of all samples during iteration, leading to time consumption.
\subsection{Algorithm Enhancement}
The GMS-RANSAC algorithm, when dealing with all matching pairs obtained from the GMS algorithm, tends to ignore the differences between feature points by randomly selecting the initial set. This oversight leads to unnecessary iterations. Since the GMS algorithm establishes a positive correlation between the neighborhood score of feature points and the probability of correct matching, it is reasonable to consider selecting the initial matching pairs based on the neighborhood score. The improved GMS-RANSAC algorithm involves sorting and grouping the matching pairs based on their neighborhood scores. The pairs with higher neighborhood scores are given priority as the initial set for solving the optimal model, aiming to reduce the number of iterations and save processing time.
\section{Experimental Results and Analysis}
In this section, experiments are conducted to validate the feasibility of the proposed method. The key aspects include the selection of evaluation criteria, choice of datasets, experimental results, parameter variation experiments, and comparative analysis with existing methods. The experiments were conducted on a laptop with an Intel i7-4710MQ processor, 8GB RAM, 500GB hard disk, and Ubuntu 18.04 operating system.
\subsection{Dataset Selection}
The experiments in this study utilize the images used in the GMS algorithm, specifically the TUM [16] and KITTI [17] datasets. These datasets provide a variety of indoor and outdoor scenes, posing significant challenges for feature extraction and matching, thereby effectively validating the proposed method. 
\subsection{Selection of Evaluation Criteria}
This study primarily employs precision, recall, and algorithm execution time as three key evaluation criteria for assessing and comparing the proposed method.

The first criterion is precision, as defined in Equation (4).
\begin{multline}
	\label{eq4}
	Precsion=(Num_{correct\_matches}\\/(Num_{correct\_matches}+Num_{false\_matches}))\times 100\%
\end{multline}
Where $Num_(correct_matches)$ is the number of correct matches, and $Num_(false_matches)$ is the number of false matches.

The second criterion is recall, as defined in Equation (5). 
\begin{multline}
	\label{eq5}
    recall=(Num_{correct\_matches}\\/(Num_{correct\_matches}+Num_{true\_matches}))\times 100\%
\end{multline}
Where $Num_(correct_matches)$ is the number of correct matches, and $Num_(true_matches)$ is the number of true matches that were not successfully matched.

The third criterion is the runtime. The computation time of the proposed algorithm includes feature point extraction time, descriptor time, and matching time, excluding input processing time and criterion calculation time.
\subsection{Experimental Results}
This section presents the experimental results of the improved GMS-RANSAC algorithm. In the algorithm flow, after extracting an equal number of ORB feature points (predefined as 3000) from two input images, a brute-force matching of feature points is performed. An example result is shown in Figure 3. 

\begin{figure}[H]
    \centering
	\begin{subfigure}{0.49\linewidth}
		\centering
		\includegraphics[width=0.99\linewidth]{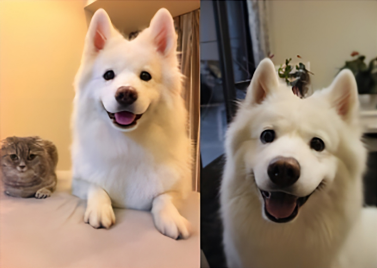}
		\caption{Input image}
		\label{Fig31}
	\end{subfigure}
	\centering
	\begin{subfigure}{0.49\linewidth}
		\centering
		\includegraphics[width=0.99\linewidth]{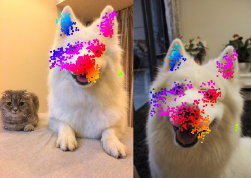}
		\caption{Result of feature point extraction}
		\label{Fig32}
	\end{subfigure}
	\centering
	\begin{subfigure}{0.49\linewidth}
		\centering
		\includegraphics[width=0.99\linewidth]{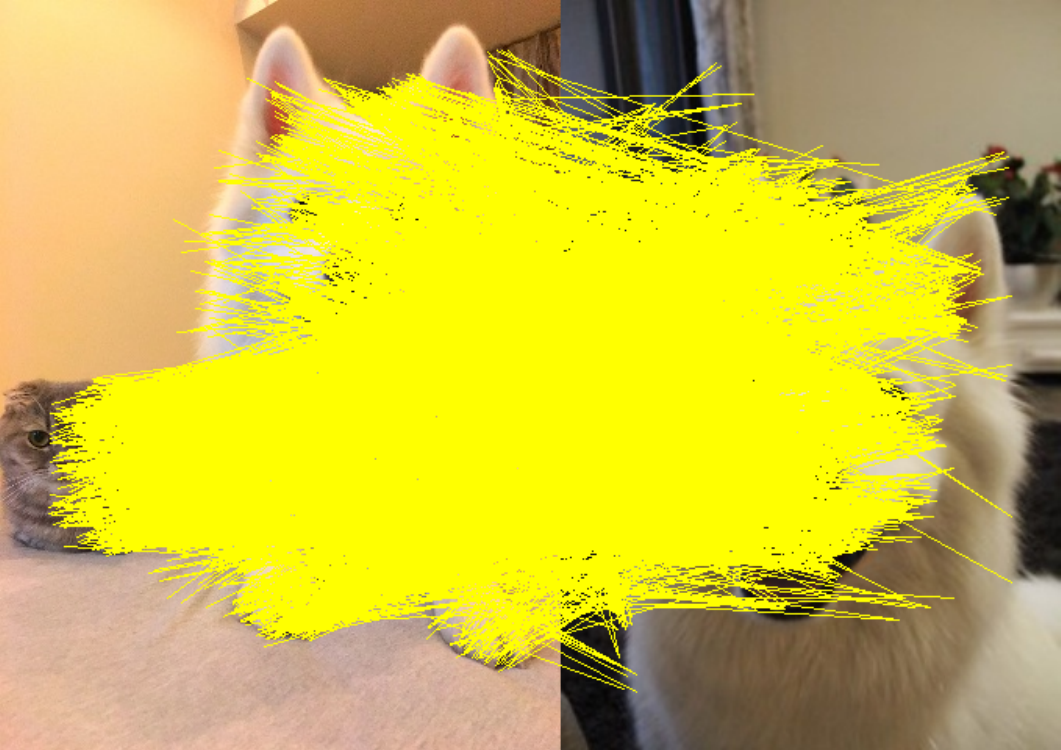}
		\caption{Result of brute-force matching}
		\label{Fig33}
	\end{subfigure}
	\caption{ Example of brute-force matching result.}
    \label{Fig3}
\end{figure}
	
Furthermore, GMS algorithm is employed for filtering, removing incorrect matches to reduce the number of erroneously matched samples input to RANSAC. The results of GMS filtering are depicted in Figure 4(a). Subsequently, RANSAC filtering is applied to the obtained results, further eliminating incorrect matches, resulting in the outcome of the proposed method, as illustrated in Figure 4(b). 
	\begin{figure}[H]
		\centering
		\begin{subfigure}{0.49\linewidth}
			\centering
			\includegraphics[width=0.99\linewidth]{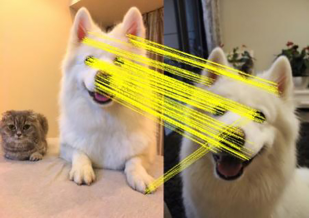}
			\caption{GMS results}
			\label{Fig41}%
		\end{subfigure}
		\centering
		\begin{subfigure}{0.49\linewidth}
			\centering
			\includegraphics[width=0.99\linewidth]{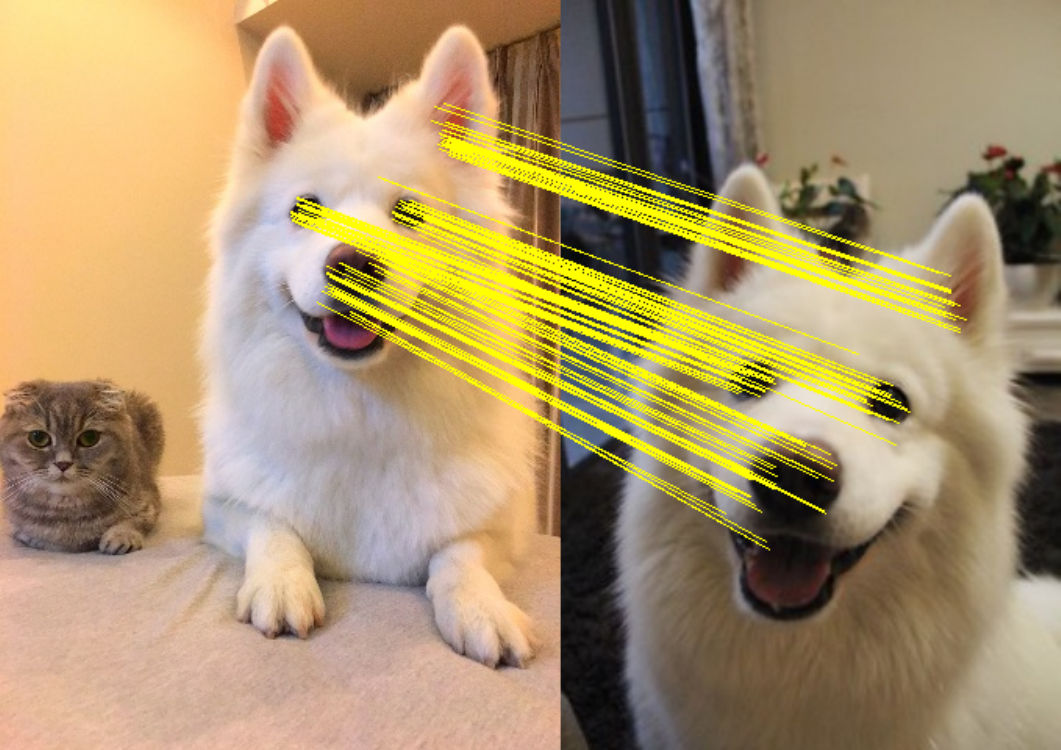}
			\caption{Improved GMS-RANSAC results}
			\label{Fig42}
		\end{subfigure}
		\caption{Screening results of different methods after brute-force matching.}
		\label{Fig4}
	\end{figure}

	\begin{figure}[H]
		\centering
		\begin{subfigure}{0.49\linewidth}
			\centering
			\includegraphics[width=0.99\linewidth]{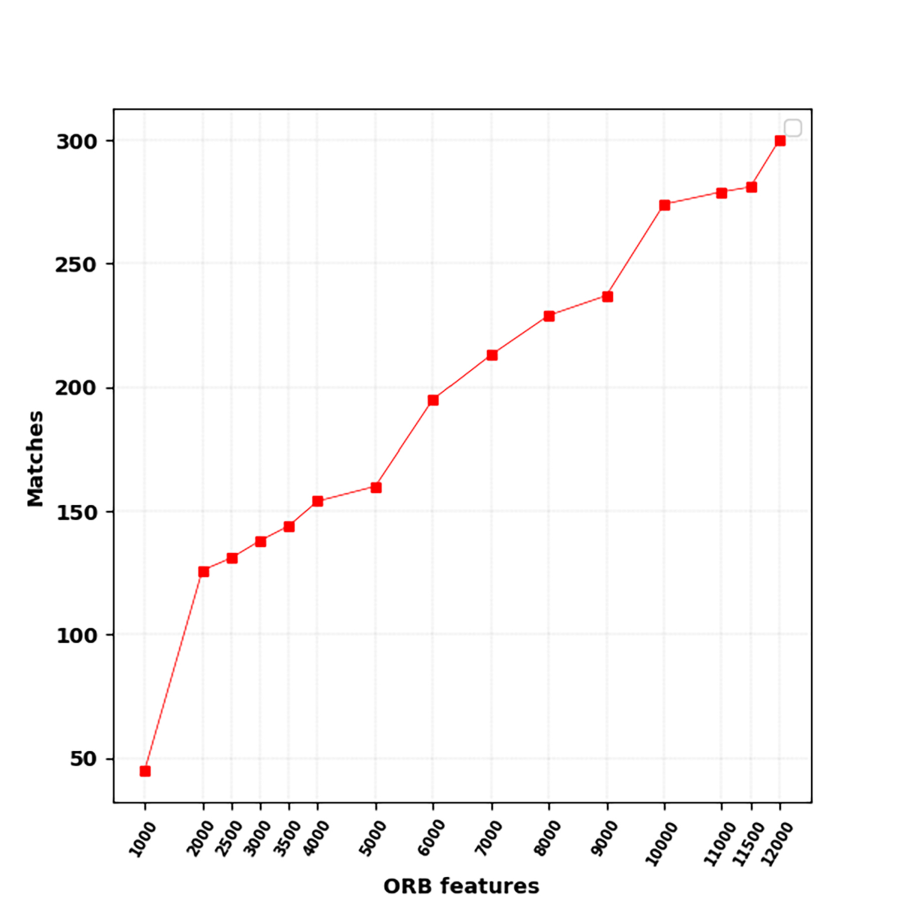}
			\caption{Matching results}
			\label{Fig51}%
		\end{subfigure}
		\centering
		\begin{subfigure}{0.49\linewidth}
			\centering
			\includegraphics[width=0.99\linewidth]{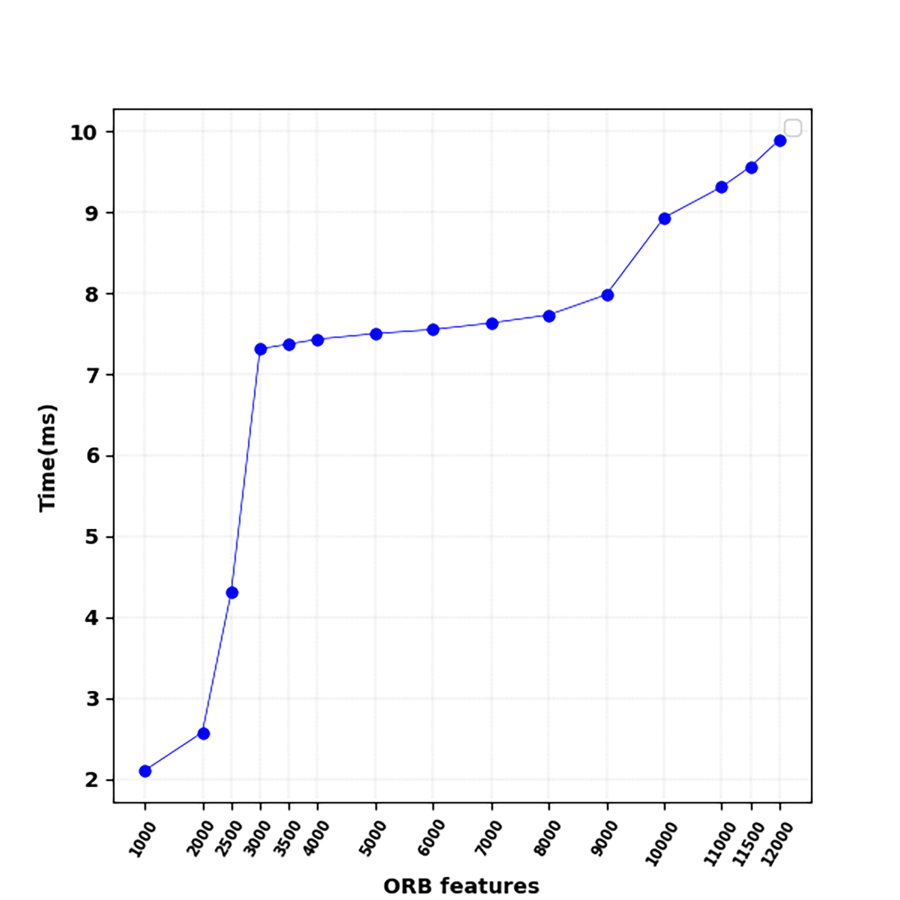}
			\caption{Time cost}
			\label{Fig52}
		\end{subfigure}
		\caption{Improved GMS-RANSAC matching results and time under different preset ORB feature point counts.}
		\label{Fig5}
	\end{figure}	
	Comparisons of the improved GMS-RANSAC algorithm under different predefined targets for the number of ORB feature points are summarized in Table 1. The experimental results are more visually represented in Figure 5. Overall, with the increase in the predefined number of feature points, both the matched pairs after eliminating incorrect matches and the processing time exhibit a sequential increase. Notably, there is a significant increase in time within the 2000-3000 range, while the increase in the number of matches is relatively modest. From 3000 onwards to 9000, the time cost remains relatively stable, and beyond 9000, there is an increasing trend in time cost. Additionally, the number of matches steadily increases and remains relatively stable.
	
	The number of extractable feature points in the images is limited, and a substantial number of matches are found before the predefined value of 2000. As the predefined value increases, further reliable matches need to be identified among the extracted feature point pairs, increasing computational workload. Consequently, while time increases, the increase in the number of matches is not as significant. In the range of 3000-9000, a small number of additional matches are extracted, and the primary increase is in search time, leading to a relatively minor increase in computation time. Beyond 9000, as the predefined value continues to rise, the number of matches found in the images decreases, and the primary increase is in search time. Therefore, the time steadily increases beyond the predefined value of 9000.
	
	It can be observed that the improved GMS-RANSAC algorithm effectively extracts ORB feature points and achieves successful matching. 
	\begin{table}[!hptb]
		\centering
		\caption{Results of Improved GMS-RANSAC under Different Predefined Numbers of ORB Feature Points.}
		\setlength{\tabcolsep}{12pt}
		\begin{tabular} {lcc}
		  \toprule
		  
		  Preset Value  & Number of Matches & Time (ms)  \\
		  \midrule
	
		  1000	&45	&2.11\\
		  2000	&126	&2.58\\
		  2500	&131	&4.31\\
		  3000	&138	&7.31\\
		  3500	&144	&7.37\\
		  4000	&154	&7.43\\
		  5000	&160	&7.50\\
		  6000	&195	&7.55\\
		  7000	&213	&7.63\\
		  8000	&229	&7.73\\
		  9000	&237	&7.98\\
		  10000	&274	&8.93\\

		  \bottomrule
		\end{tabular}
		
		\label{tab1}
	  \end{table}   

	  \subsection{Parameter Variation Experiment}
	  In this section, we will demonstrate the impact of key parameters on the number of matches and processing speed, enabling us to identify the optimal parameter values. The key parameter in our method is the proportion of grouping after sorting by confidence. We test the influence of different proportions on the number of matches and speed using the same images as in the previous section, under various predefined values for the number of ORB feature points. The results are presented in Figure 6, Table 2, and Table 3. 
    
	\begin{figure}[H]
		\centering
		\begin{subfigure}{0.49\linewidth}
			\centering
			\includegraphics[width=0.99\linewidth]{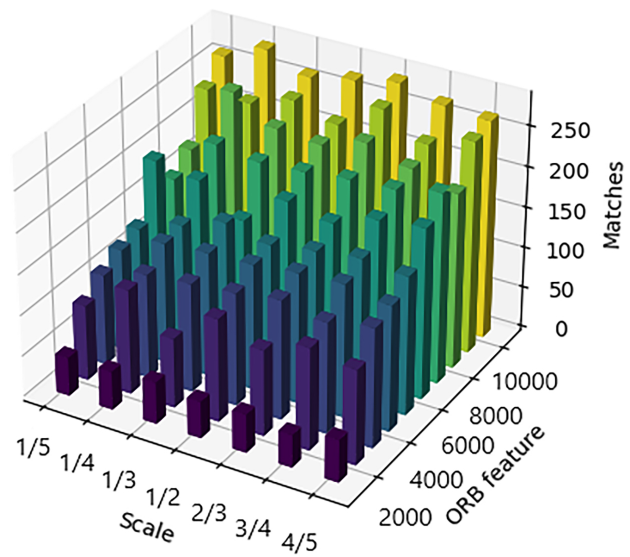}
			\caption{Matching results}
			\label{Fig61}%
		\end{subfigure}
		\centering
		\begin{subfigure}{0.49\linewidth}
			\centering
			\includegraphics[width=0.99\linewidth]{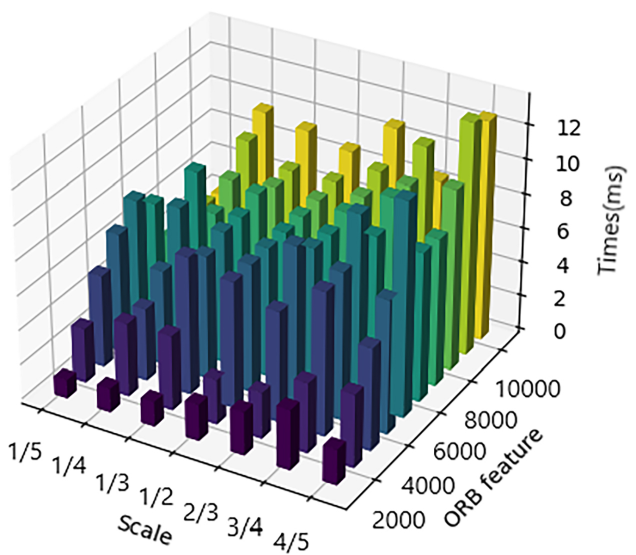}
			\caption{Time cost}
			\label{Fig62}
		\end{subfigure}
		\caption{Trend graph of matching results for different grouping ratios under different preset numbers of feature points.}
		\label{Fig6}
	\end{figure}

	\begin{table}[!hptb]
		\centering
		\caption{Number of matches at different grouping ratios under different preset numbers of feature points.}
		\setlength{\tabcolsep}{7pt}
		\begin{tabular} {lccccccc}
		  \toprule
		  
		  Preset Value & 1/5	&1/4	&1/3	&1/2	&2/3	&3/4	&4/5 \\
		  \midrule
		  1000	&47		&47		&51 	&45		&47		&40		&52\\
		  2000	&92		&128	&85		&126	&105	&125	&116\\
		  3000	&111	&128	&133	&138	&146	&136	&146\\
		  4000	&124	&148	&152	&154	&160	&163	&154\\
		  5000	&132	&152	&171	&160	&168	&175	&170\\
		  6000	&199	&190	&156	&195	&184	&203	&209\\
		  7000	&159	&217	&210	&213	&219	&222	&233\\
		  8000	&179	&264	&234	&229	&246	&230	&215\\
		  9000	&231	&234	&252	&237	&270	&242	&260\\
		  10000	&262	&284	&264	&274	&285	&272	&268\\

		  \bottomrule
		\end{tabular}
		
		\label{tab2}
	  \end{table} 
	  
	  \begin{table}[!hptb]
		\centering
		\caption{Time required to complete matching at different grouping ratios under different preset numbers of feature points (ms).}
		\setlength{\tabcolsep}{5pt}
		\begin{tabular} {lccccccc}
		  \toprule
		  
		  Preset Value & 1/5	&1/4	&1/3	&1/2	&2/3	&3/4	&4/5 \\
		  \midrule
		1000	&1.06	&1.38	&1.44	&2.11	&2.52	&3.48	&2.05   \\
		2000	&3.13	&4.23	&4.31	&2.58	&2.67	&4.06	&4.23\\
		3000	&5.34	&4.18	&7.95	&7.31	&6.39	&8.29	&5.89\\
		4000	&6.91	&5.5	&7.13	&7.43	&9.23	&8.5	&7.69\\
		5000	&8.03	&8.34	&7.68	&7.5	&8.24	&10.91	&12.38\\
		6000	&7.01	&9.56	&7.7	&7.55	&8.17	&8.81	&8.62\\
		7000	&4.2	&6.33	&8.22	&7.63	&8.58	&10.16	&8.54\\
		8000	&4.64	&7.46	&7.79	&7.73	&8.65	&10.01	&10.5\\
		9000	&4.54	&8.96	&7.91	&7.98	&9.28	&11.41	&13.46\\
		10000	&4.23	&9.87	&9.46	&8.93	&10.97	&8.62	&12.76\\

		  \bottomrule
		\end{tabular}
		
		\label{tab3}
	  \end{table} 
	  Combining the data from Figure 6 and Tables 2 and 3, it can be observed that under different predefined values, the overall trend of time varies with the increase in grouping proportion, showing an initial increase, followed by a decrease, and then an increase. However, there are differences in the trend at different predefined values. While the number of matches follows the trend of increasing with the predefined number of feature points, the impact of grouping proportion shows different trends under different predefined values.
	  
	  In the 1000-2000 range, the number of matches exhibits a continuous increase and then decrease trend with the increase in grouping proportion. In terms of time, there is an upward trend on both sides of 1/2 as the dividing point. In the 3000-4000 range, the time with grouping proportions of 1/5 and 1/4 is significantly less than other proportions, but the number of matches is also lower. The time for other proportions is relatively close, and the number of matches increases with the proportion. At the predefined value of 5000, the time is minimal when the grouping proportion is 1/2, and the number of matches is higher than that of 1/5 and 1/4 groupings. In the 6000-10000 range, the time is minimal when the grouping proportion is 1/5, and only at the predefined value of 6000, the number of matches is higher than other proportions. At 1/2 grouping proportion, the time is less, and the number of matches is higher than the average for other proportions.
	  
	  In visual SLAM, when dealing with large datasets, the time spent on feature matching is crucial for the real-time performance of the entire SLAM system. The aim of this study is to reduce the time expenditure while ensuring sufficient accuracy in feature matching. Based on the above analysis, a grouping proportion of 1/2 can meet our requirements. Therefore, this study selects a grouping proportion of 1/2 for further experiments.
	  \subsection{ Comparative Analysis of Methods}
	  In this section, a comparative analysis between the proposed method and other similar methods is presented to highlight the advantages of the proposed approach. In the experiments, comparisons are made between GMS-RANSAC [7], an improved GMS-RANSAC algorithm proposed by Lan et al. [18], and our improved GMS-RANSAC method on the TUM desk, KITTI, and TUM doll datasets. The GMS and RANSAC components in all three methods use the same parameters and thresholds where there is no improvement.
	  
	  Figure 7 illustrates the comparison results between the GMS-RANSAC algorithm and our proposed improved GMS-RANSAC algorithm when extracting 3000 ORB feature points from different datasets. In (a), (b), and (c), the top images represent the results of the GMS-RANSAC algorithm on the TUM and KITTI datasets, while the others depict the results of our proposed improved GMS-RANSAC algorithm. From the figures, it is evident that our proposed method yields more accurate matching results compared to the GMS-RANSAC method.
	  
	  \begin{figure}[!hptb]	
		\centering
		\begin{subfigure}{0.45\linewidth}
			\centering
			\includegraphics[width=0.95\linewidth]{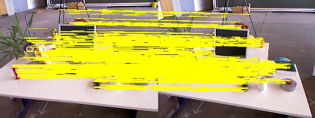}
			\caption{TUM desk dataset}
			\label{Fig71}%
		\end{subfigure}
		\centering
		\begin{subfigure}{0.45\linewidth}
			\centering
			\includegraphics[width=0.95\linewidth]{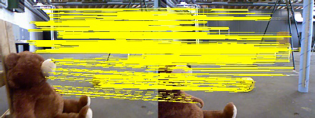}
			\caption{TUM doll dataset}
			\label{Fig112}
		\end{subfigure}
		\centering
		\begin{subfigure}{0.45\linewidth}
			\centering
			\includegraphics[width=0.95\linewidth]{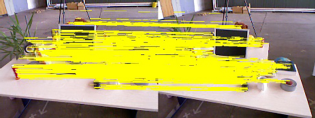}
			\caption{TUM desk dataset}
			\label{Fig113}%
		\end{subfigure}
		\centering
		\begin{subfigure}{0.45\linewidth}
			\centering
			\includegraphics[width=0.95\linewidth]{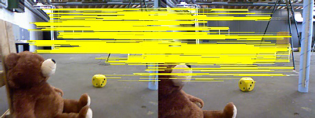}
			\caption{TUM doll dataset}
			\label{Fig114}
		\end{subfigure}
		\centering
		\begin{subfigure}{0.95\linewidth}
			\centering
			\includegraphics[width=0.95\linewidth]{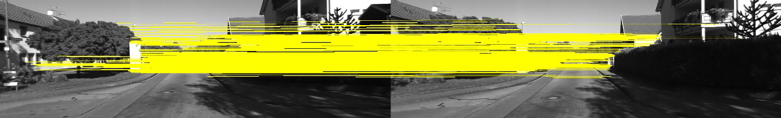}
			\caption{KITTI dataset}
			\label{Fig115}%
		\end{subfigure}
		\centering
		\begin{subfigure}{0.95\linewidth}
			\centering
			\includegraphics[width=0.95\linewidth]{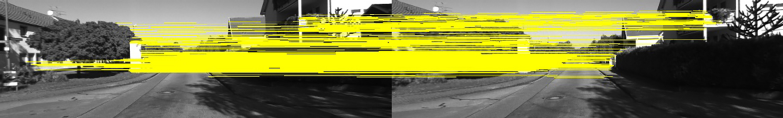}
			\caption{KITTI dataset}
			\label{Fig76}
		\end{subfigure}
		\caption{Comparison between GMS-RANSAC and our proposed improved GMS-RANSAC when extracting 3000 ORB feature points from different datasets. The top row represents GMS-RANSAC, and the bottom row represents our proposed method.}
	\end{figure}
	To further validate the effectiveness of our method, we evaluate our approach against the GMS-RANSAC algorithm and an improved GMS-RANSAC algorithm proposed in reference [18], by comparing the matching accuracy and the time spent on matching. Table 4 presents the precision and recall of the three methods when extracting 2000 ORB feature points for matching on three datasets. Our method performs best on the TUM desk dataset, while the GMS-RANSAC algorithm and the improved GMS-RANSAC algorithm from reference [18] excel on the KITTI and TUM doll datasets. However, on datasets where our method does not outperform the other methods, the precision and recall are comparable, indicating similar capabilities in eliminating incorrect matches.
	
	To further confirm that our method, the GMS-RANSAC algorithm, and the improved GMS-RANSAC algorithm from reference [18] exhibit similar matching accuracy, we conducted a comparative analysis by extracting different predefined numbers of ORB feature points on three datasets. The number of matches serves as a measure of matching accuracy. Figure 8 demonstrates that as the predefined number of ORB feature points increases, the number of matches obtained by all three methods gradually increases. Between predefined values of 1000 and 4000, the number of matches for all three methods is nearly identical. In the range of predefined values from 5000 to 10000, there is a slight difference in the number of matches among the three methods, but the difference is minimal. Thus, our proposed improved GMS-RANSAC algorithm demonstrates matching accuracy on par with the GMS-RANSAC algorithm and the improved GMS-RANSAC algorithm from reference [18].
	
	\begin{table}[!hptb]
		\centering
		\caption{Precision and Recall of Different Methods on Three Datasets Extracting 2000 ORB Feature Points for Matching.}
		\setlength{\tabcolsep}{4.5pt}
		\begin{tabular} {lcccc}
		  \toprule
		  
		  Dataset &  & GMS-RANSAC & [18] & OURS  \\
		  \midrule

		  \multirow{2}{*}{TUM desk}	&Precision\%	&75.5501	&76.8519	&77.0202\\
		&Recall Rate\%	&44.85	&45.65	&45.75\\

		\multirow{2}{*}{KITTI}	&Precision\%	&65.6287	&55.4591	&61.5569\\
		&Recall Rate\%	&27.4	&23.15	&25.7\\

		\multirow{2}{*}{TUM doll}	&Precision\%	&81.9466	&77.9435	&77.551\\
		&Recall Rate\%	&52.2	&4965	&49.4\\

		  \bottomrule
		\end{tabular}
		
		\label{tab4}
	  \end{table}  

	  \begin{figure}[H]
		\centering
		\begin{subfigure}{0.45\linewidth}
			\centering
			\includegraphics[width=0.95\linewidth]{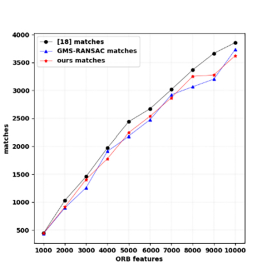}
			\caption{TUM desk dataset}
			\label{Fig81}
		\end{subfigure}
		\centering
		\begin{subfigure}{0.45\linewidth}
			\centering
			\includegraphics[width=0.95\linewidth]{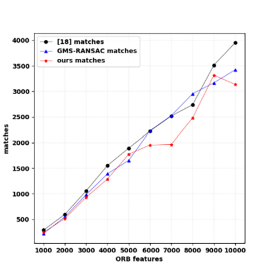}
			\caption{KITTI dataset}
			\label{Fig82}
		\end{subfigure}
		\centering
		\begin{subfigure}{0.45\linewidth}
			\centering
			\includegraphics[width=0.95\linewidth]{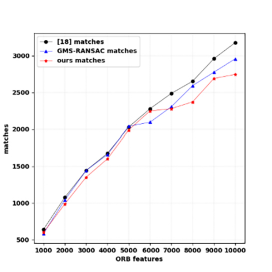}
			\caption{TUM doll dataset }
			\label{Fig83}
		\end{subfigure}
		\caption{Example of matching results.}
		\label{Fig8}
	\end{figure}

	\begin{figure}[H]
		\centering
		\begin{subfigure}{0.45\linewidth}
			\centering
			\includegraphics[width=0.95\linewidth]{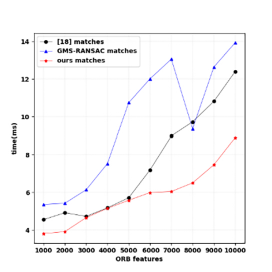}
			\caption{TUM desk dataset}
			\label{Fig91}
		\end{subfigure}
		\centering
		\begin{subfigure}{0.45\linewidth}
			\centering
			\includegraphics[width=0.95\linewidth]{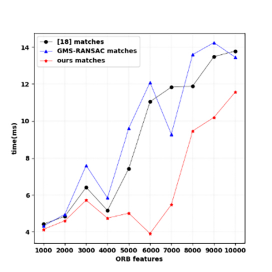}
			\caption{KITTI dataset}
			\label{Fig92}
		\end{subfigure}
		\centering
		\begin{subfigure}{0.45\linewidth}
			\centering
			\includegraphics[width=0.95\linewidth]{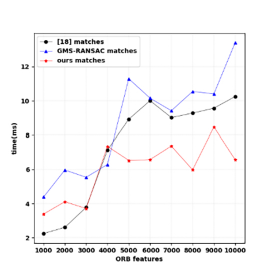}
			\caption{TUM doll dataset }
			\label{Fig93}
		\end{subfigure}
		\caption{Time cost of three algorithms on different datasets.}
		\label{Fig9}
	\end{figure}
	With similar accuracy, we validate the speed advantage of our method by comparing the time costs of three approaches. Figure 9 displays the time costs of the three methods on the TUM and KITTI datasets, gradually increasing with the predefined number of ORB feature points. It is evident from the figure that as the number of ORB feature points increases, the time gap between the three methods gradually widens, and our method consistently exhibits the lowest time costs across all three datasets.
	
	To more intuitively showcase the time advantage of our proposed method over the GMS-RANSAC algorithm and the improved GMS-RANSAC algorithm from reference \cite{ref18}, we computed the average time differences between our proposed method and the other two methods at different predefined values, as shown in Table 5. Under the same conditions, our method achieves a precision improvement of 1.4701\% and 0.1683\% on the TUM desk dataset while reducing the time consumption by 37.19\% and 18.26\% compared to the GMS-RANSAC algorithm and the algorithm from reference \cite{ref18}, respectively. On the KITTI and TUM doll datasets, maintaining similar precision to the GMS-RANSAC algorithm and the algorithm from reference \cite{ref18}, our method reduces the runtime by 28.32\% and 30.48\% compared to the GMS-RANSAC algorithm, and by 24.22\% and 6.29\% compared to the algorithm from reference \cite{ref18}. Across all three datasets, our proposed improved GMS-RANSAC algorithm achieves an average time reduction of 32\% compared to the GMS-RANSAC algorithm and 16.26\% compared to the algorithm from reference \cite{ref18}. Therefore, our proposed improved GMS-RANSAC algorithm, while ensuring matching accuracy, exhibits a significant speed advantage over the GMS-RANSAC algorithm and the algorithm from reference \cite{ref18} with an average time reduction of 24.13\% across the TUM desk, TUM doll, and KITTI datasets.
	\begin{table}[!hptb]
		\newcommand{\tabincell}[2]{\begin{tabular}{@{}#1@{}}#2\end{tabular}}
		\centering
		\caption{Average Time Differences between Our Proposed Method and Other Methods at Different Predefined Values.}
		\setlength{\tabcolsep}{4pt}
		\begin{tabular} {lccc}
		  \toprule
		  
		  Dataset &  \tabincell{c}{Time Reduction Compared \\to GMS-RANSAC} & \tabincell{c}{Time Reduction \\Compared to [18] } \\
		  \midrule
	
		  TUM desk[16]	&37.19\%	&18.26\%\\
		  KITTI[17]	&28.32\%	&24.22\%\\
		  TUM doll[16]	&30.48\%	&6.29\%\\
		  Average Value	&32.00\%	&16.26\%\\

		  \bottomrule
		\end{tabular}
		
		\label{tab5}
	  \end{table}

  \section{Conclusion}
  This study proposes an improved fast elimination method for incorrect matches based on the GMS-RANSAC algorithm. The method demonstrates reduced processing time for scenarios with either too many or too few ORB feature points compared to the existing GMS-RANSAC method. Experimental results indicate that our approach not only improves precision on the TUM desk dataset while reducing time consumption but also ensures comparable accuracy to GMS-RANSAC on the KITTI and TUM doll datasets with shortened runtime. Consequently, this method, while maintaining accuracy similar to the GMS-RANSAC algorithm, reduces average runtime by 24.13\% compared to the runtime of the GMS-RANSAC algorithm and the improved algorithm from reference \cite{ref18} across the KITTI, TUM desk, and TUM doll datasets.





\end{document}